\documentclass[sigconf]{acmart}
\AtBeginDocument{%
  \providecommand\BibTeX{{%
    \normalfont B\kern-0.5em{\scshape i\kern-0.25em b}\kern-0.8em\TeX}}}

\usepackage{booktabs}
\usepackage{multirow}
\usepackage{array}

\copyrightyear{2021}
\acmYear{2021}
\setcopyright{acmcopyright}\acmConference[SIGGRAPH '21 Emerging Technologies ]{Special Interest Group on Computer Graphics and Interactive Techniques Conference Emerging Technologies }{August 9--13, 2021}{Virtual Event, USA}
\acmBooktitle{Special Interest Group on Computer Graphics and Interactive Techniques Conference Emerging Technologies (SIGGRAPH '21 Emerging Technologies ), August 9--13, 2021, Virtual Event, USA}
\acmPrice{15.00}
\acmDOI{10.1145/3450550.3465349}
\acmISBN{978-1-4503-8364-6/21/08}

\begin{document}

\title{DronePaint: Swarm Light Painting with DNN-based Gesture Recognition
}

\author{Valerii Serpiva}
\affiliation{%
  \institution{Skolkovo Institute of Science and Technology}
  \streetaddress{Nobel street 3}
  \city{Moscow}
  \country{Russia}}
\email{valerii.serpiva@skoltech.ru}

\author{Ekaterina Karmanova}
\affiliation{%
  \institution{Skolkovo Institute of Science and Technology}
  \streetaddress{Nobel street 3}
  \city{Moscow}
  \country{Russia}}
\email{ekaterina.karmanova@skoltech.ru}

\author{Aleksey Fedoseev}
\affiliation{%
  \institution{Skolkovo Institute of Science and Technology}
  \streetaddress{Nobel street 3}
  \city{Moscow}
  \country{Russia}}
\email{aleksey.fedoseev@skoltech.ru}

\author{Stepan Perminov}
\affiliation{%
  \institution{Skolkovo Institute of Science and Technology}
  \streetaddress{Nobel street 3}
  \city{Moscow}
  \country{Russia}}
\email{stepan.perminov@skoltech.ru}

\author{Dzmitry Tsetserukou}
\affiliation{%
  \institution{Skolkovo Institute of Science and Technology}
  \streetaddress{Nobel street 3}
  \city{Moscow}
  \country{Russia}}
\email{d.tsetserukou@skoltech.ru}

\renewcommand{\shortauthors}{Serpiva, Karmanova, Fedoseev, Perminov and Tsetserukou}

\begin{abstract}


We propose a novel human-swarm interaction system, allowing the user to directly control a swarm of drones in a complex environment through trajectory drawing with a hand gesture interface based on the DNN-based gesture recognition. 

The developed CV-based system allows the user to control the swarm behavior without additional devices through human gestures and motions in real-time, providing convenient tools to change the swarm’s shape and formation. The two types of interaction were proposed and implemented to adjust the swarm hierarchy: trajectory drawing and free-form  trajectory  generation  control.

The experimental results revealed a high accuracy of the gesture recognition system (99.75\%), allowing the user to achieve relatively high precision of the trajectory drawing (mean error of 5.6 cm in comparison to 3.1 cm by mouse drawing) over the three evaluated trajectory patterns.
The proposed system can be potentially applied in complex environment exploration, spray painting using drones, and interactive drone shows, allowing users to create their own art objects by drone swarms.

\end{abstract}



\begin{teaserfigure}
  \includegraphics[width=\textwidth]{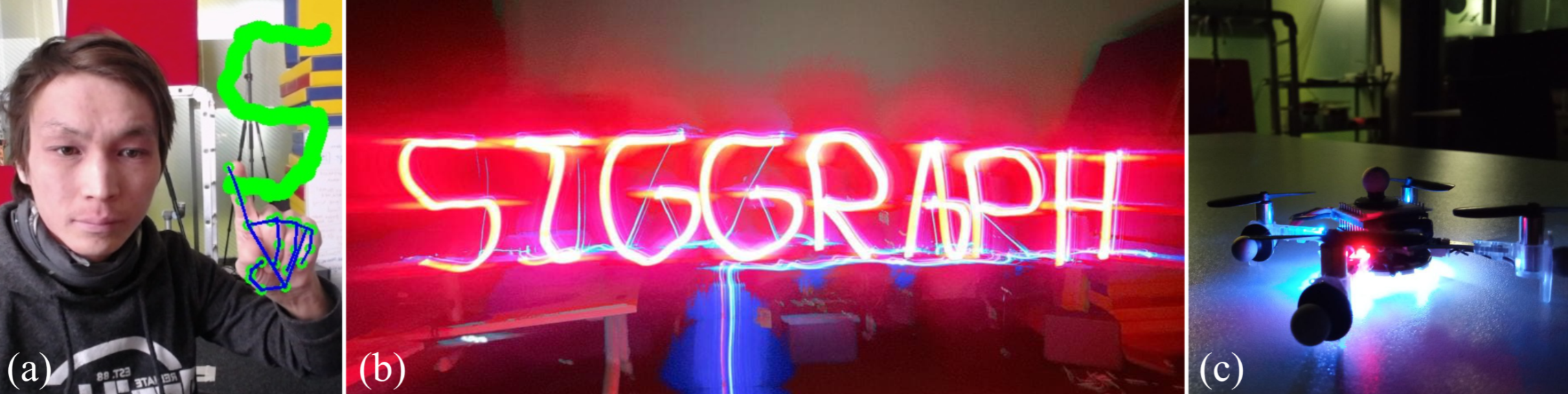}
  \caption{a) View of user screen. Drawing trajectory by hand and gesture recognition. b) Long exposure light painting of “Siggraph" logo by drone. c) Single swarm agent with the LED circle.}
  \label{fig:teaser}
\end{teaserfigure}

\maketitle

\section{Introduction}




Human and aerial swarm interaction (HSI) nowadays serves multiple purposes, such as search and rescue operations, cargo delivery, remote inspection, etc. One of the prominent implementations of the robotic swarms had recently emerged in art industry, where drones perform as scalable and interactive tools for light and spray painting. For example, an autonomous drone equipped with a spray gun-holding arm was developed by \cite{Vempati_2018} for spray painting on various three-dimensional surfaces. Furthermore, a multi-drone graffiti was proposed by \cite{Uryasheva_2019} with a task dispatch system based on the parametric greedy algorithm.

Several research papers focus on interactive art concepts where drones provide the color pallet by LED arrays. For instance, \cite{Dubois_2015} proposed an interactive choreographic show where humans and drones move synchronously with precise swarm behavior computation. A tangible experience of HSI was introduced by \cite{Gomes_2016} where drones serve as a colorful interactive 3D display. Another practical approach was presented by \cite{Knierim_2018}, who proposed drones with light beacons as a navigation system that projects the map instructions into the real world.

With these considerations, a real-time control interface over the swarm is required to deliver the user an immersive real-time experience of painting. Many researchers propose gesture-based interfaces as a versatile and intuitive tool of HSI. For example, a tactile interface for HSI with an impedance-based swarm control was developed by \cite{Tsykunov_2019}. A multi-channel robotic system for HSI in augmented reality was suggested by \cite{Chen_2020}. \cite{Suresh_2019} proposed a complex control approach with arm gestures and motions, which are recorded by a wearable armband, controlling a swarm’s shape and formation. \cite{Alonso-Mora_2015} and \cite{Kim_2020} suggested real-time input interfaces with swarm formation control. However, their approach was developed only for mobile robot operation in 2D space. The wearable devices for high mobility of the user were proposed by \cite{Byun_2019}, suggesting an epidermal tactile sensor array to achieve the direct teleoperation of the swarm by human hand.

Previously developed systems have achieved low time delays and high precision of the control, yet their trajectory generating capability is limited to the direct gesture input and simple hand motions. We propose the DronePaint system for drone light painting with DNN-based gesture recognition to make the way we communicate with drones intuitive and intelligent. Only with a single camera and developed software any not-experienced user will be capable of generating impressive light drawings in midair. 



\section{System overview}
\begin{figure}[!ht]
    \centering
    \includegraphics[width=0.95\linewidth]{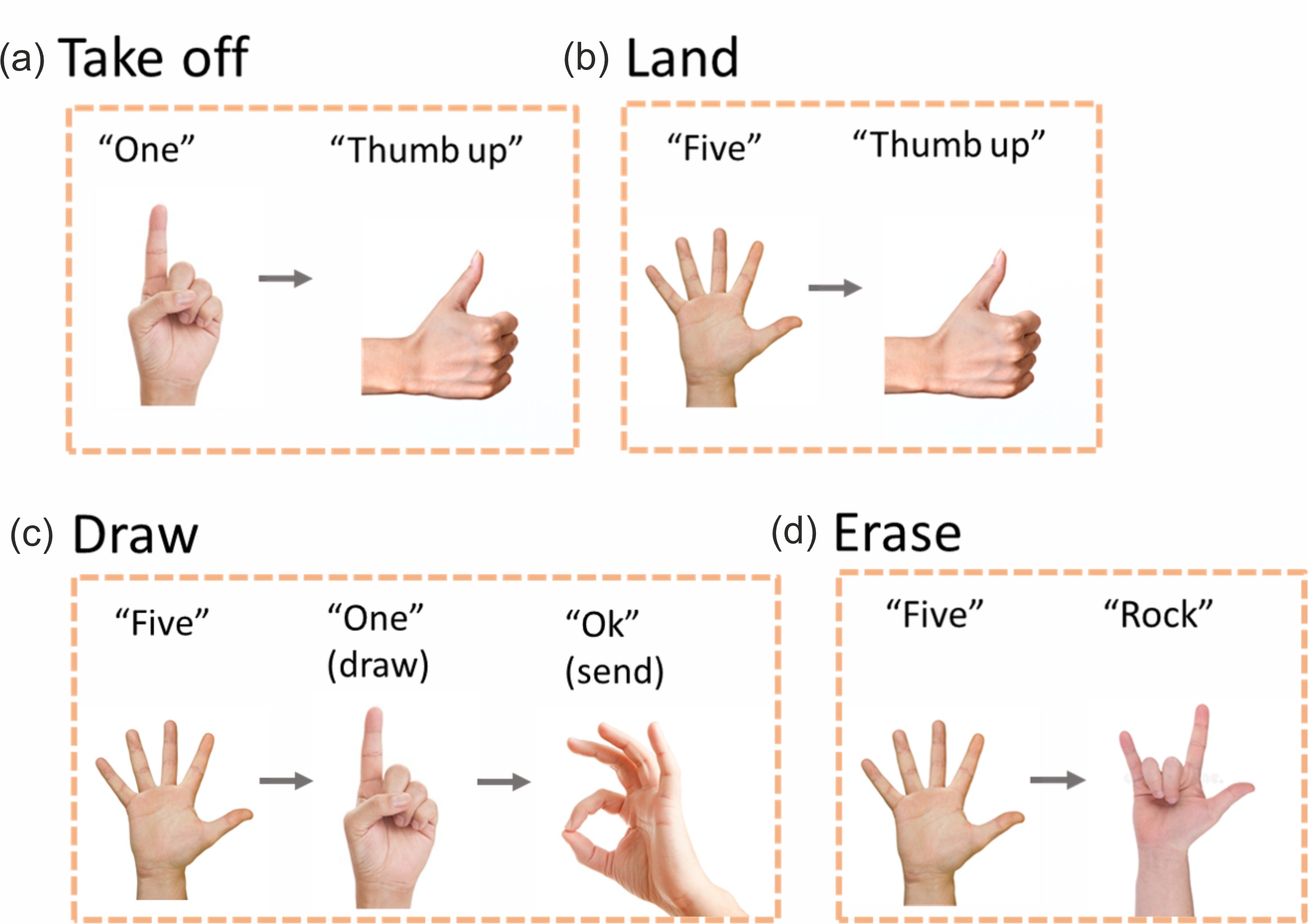}
    \caption{Example of gestures and commands to control a drone.}
    \label{fig:commands}
\end{figure}

Before deploying the swarm of drones, the operator positions themselves in front of the webcam, which sends the captured footage to the gesture recognition module. As soon as DronePaint is activated, the module starts to recognize the operator's position and hand gestures, awaiting the “take off" command to deploy the swarm (Fig. \ref{fig:commands}a). After the drones take off, the gesture recognition module generates a trajectory drawn by the operator with gestures. The developed DronePaint interface allows the operator both to draw and erase the trajectory (Fig. \ref{fig:commands}c, d) to achieve the desired result. After that, the trajectory is processed by the trajectory planning module to make it suitable for the swarm of drones. Then, the processed trajectory is sent simultaneously to the swarm control module and the flight simulation module using the ROS framework. Finally, the swarm of drones is led by the swarm control module by the received trajectory.

Drones complete their work at the command “land" performed by the operator with the appropriate gesture (Fig. \ref{fig:commands}b).

\subsection{System architecture}
The developed DronePaint system software consists of three modules: human-swarm interface, trajectory processing module, and drone control system.

\begin{figure}[htbp]
    \centering
    \includegraphics[width=0.95\linewidth]{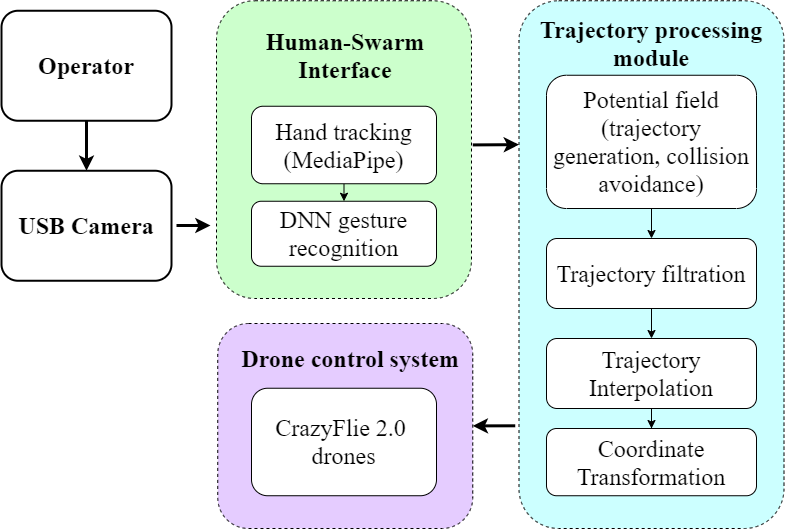}
    \caption{System architecture. The three key modules: human-swarm interface, trajectory processing module, and drone control system.}
    \label{fig:node_graph}
\end{figure}

The hardware part consists of Vicon Tracking system with 12 IR cameras for drone positioning and PC with mocap framework, a PC with CV system and drone-control framework, Logitech HDPro Webcam C920 of @30FPS for recognizing the operator hand movements and gestures, small quadcopters Crazyflie 2.0, and PC with Unity environment for visual feedback provided to the operator (Fig. \ref{fig:node_graph}). 
Communication between all systems is performed by ROS framework.

\subsection{Gesture recognition}

The human-swarm interface consists of two components: hand tracking and gesture recognition modules. The hand tracking module is implemented on the base of the Mediapipe framework. It provides high-fidelity tracking of the hand by employing Machine Learning (ML) to infer 21 key points of a human hand per a single captured frame. The gesture recognition module is based on Deep Neutral Network (DNN) to achieve high precision in human gesture classification, used for drone control and trajectory generation.

\begin{figure}[!ht]
    \centering
    \includegraphics[width=0.9\linewidth]{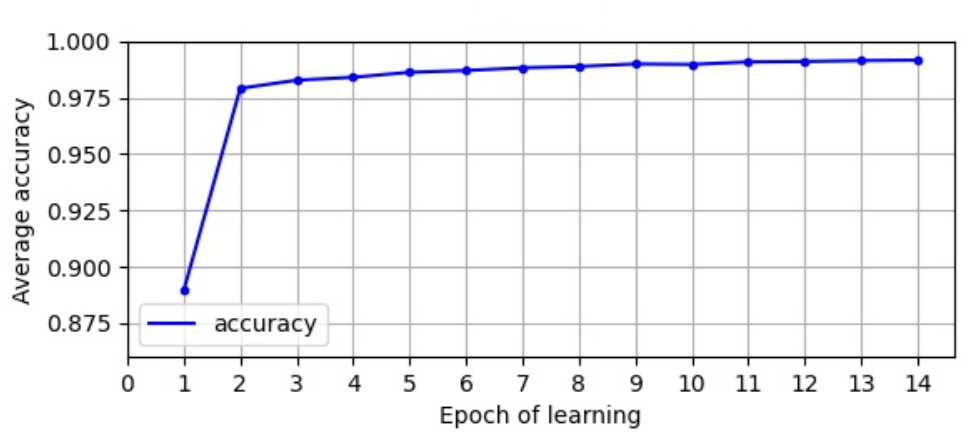}
    \caption{The classification accuracy of the developed gesture recognition system.}
    \label{fig:acc_loss}
\end{figure}

For convenient drone control we propose 8 gestures: “one", “two", “three", “four", “five", “okay", “rock", and “thumbs up". A gesture dataset for the model training was recorded by five participants. It consists of 8000 arrays with coordinates of 21 key points of a human hand: 1000 per each gesture (200 per each person). We used normalized landmarks, i.e., angles between joints and pairwise landmark distances as features to predict the gesture class.
It resulted in accuracy of 99.75\% when performing validation on a test set (Fig. \ref{fig:acc_loss}).

Using coordinates of the landmarks, we calculate the hand coordinates and size on the image received from the USB camera. To calculate the distance between the hand and camera the value of the palm size was applied.


\subsection{Trajectory processing}

\begin{figure}[!ht]
    \centering
    \includegraphics[width=0.9\linewidth]{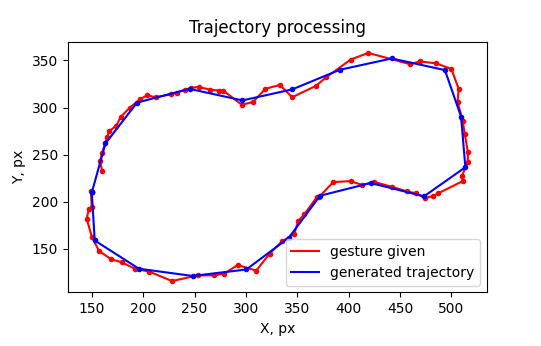}
    \caption{Hand trajectory normalization. Trajectory recorded from hand movement (red line). Trajectory with filtration and linear interpolation (blue line).}
    \label{fig:traj_processing}
\end{figure}

To ensure a smooth flight of the drone, a trajectory with equidistant flight coordinates is required. The drawn trajectory may be uneven or contain unevenly distributed coordinates. Therefore, the trajectory processing module smooths the drawn trajectory with an alpha-beta filter (filtration coefficient equals 0.7) and then interpolates it uniformly (Fig. \ref{fig:traj_processing}). After that, the trajectory coordinates are transformed from the DronePaint interface screen (pixels) to the flight zone coordinate system (meters). 
The coordinates of the generated trajectory are sequentially transferred to the drone control system using ROS framework. The time intervals between each coordinate transfer depend on the distance between the coordinates and the flight speed of the drone.

\subsection{Swarm control algorithm}

The potential field approach was adjusted and applied to UAVs for robust path planning and collision avoidance between the swarm units. The basic principle of this method lies in the modeled force field, which is composed of two opposing forces, i.e., attractive force and repulsive force.
The attractive force pulls the UAV to the desired position, located on the drawn and processed trajectory, while the repulsive force repels the UAV. The repulsive force centers are located on the obstacle surfaces and the other UAVs.




\section{Experimental Evaluation}
\subsubsection*{Procedure}
\begin{figure}[ht]
    \centering
    \includegraphics[width=0.7\linewidth]{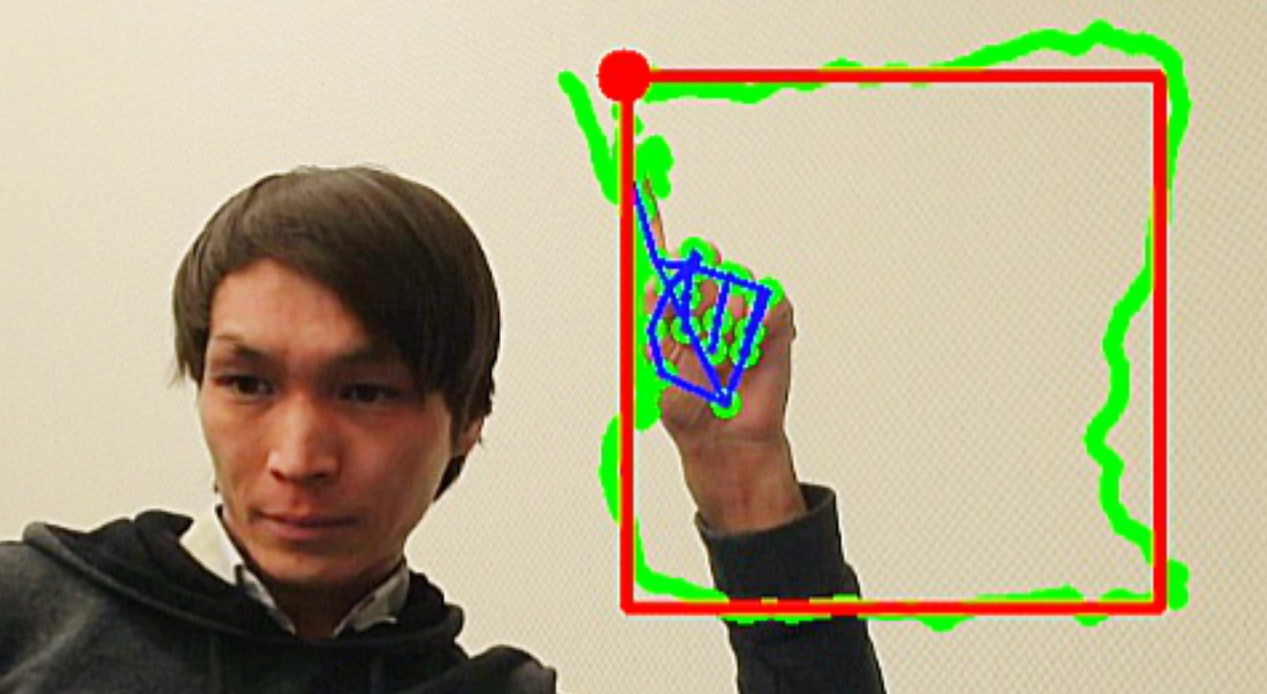}
    \caption{View of user screen. Hand-drawn trajectory and target trajectory of square shape are illustrated by red and green lines, respectively.}
    \label{fig:drawing_process}
\end{figure}

We invited 7 participants aged 22 to 28 years (mean=24.7, std=1.98) to test DronePaint system. 14.3\% of them have never interacted with drones before, 28.6\% regularly deal with drones, almost 87\% of participants were familiar with CV-based systems or had some experience with gesture recognition. To evaluate the performance of the proposed interface, we collected trajectories drawn by gestures (Fig. \ref{fig:drawing_process}) and computer mouse from the participants.
Drawing by the mouse is a reference point in determining the convenience and accuracy of the proposed method.


\begin{figure}
    \includegraphics[width=1.0\linewidth]{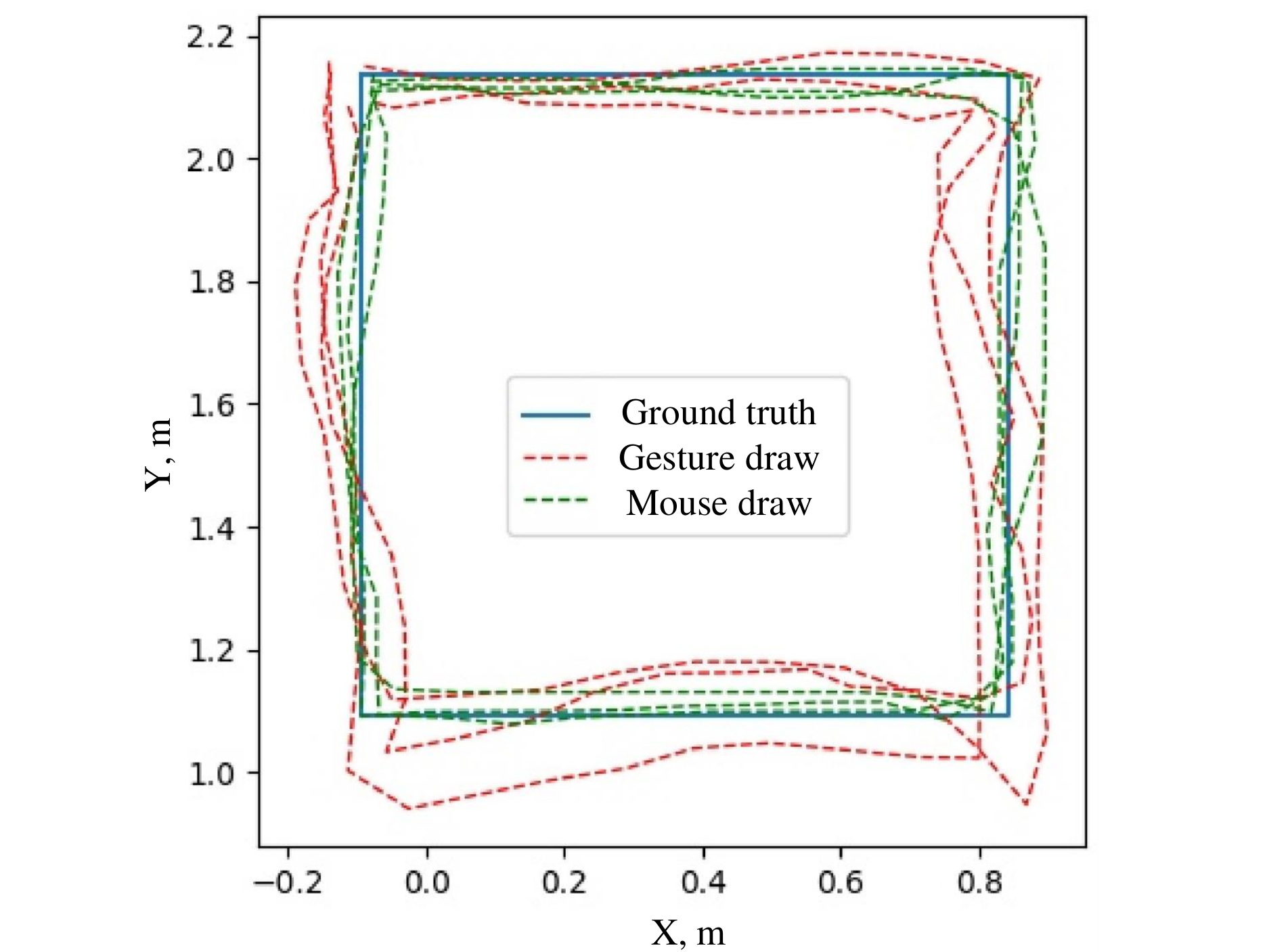}
\caption{Square trajectory drawn by the hand gestures (red dashed line) and mouse (green dashed line).}
\label{fig:plots}
\end{figure}

After that, the three best attempts were chosen to evaluate the performance of users with two trajectory generating interfaces. Fig. \ref{fig:plots} shows two of three ground truth paths (solid blue line), where for each one a user traces the path several times by hand gesture motion (red dashed line) and a mouse (green dashed line).

\subsection{Trajectory tracing error}
The comparative results of gesture-drawn and mouse-drawn trajectory generation are presented in Table \ref{tab:error}. 

\begin{table}[ht]
\caption{Comparison experiment for trajectories drawn by the hand (H) and mouse (M).}
\begin{tabular}{|c|c|c|c|c|c|c|}
\hline
\multirow{3}{*}{} & \multicolumn{6}{c|}{Trajectories}                  \\ \cline{2-7} 
                  & \multicolumn{2}{c|}{Square} & \multicolumn{2}{c|}{Circle} & \multicolumn{2}{c|}{Triangle} \\ \cline{2-7} 
                  & H             & M           & H             & M           & H              & M            \\ \hline
Max error, cm     & 18.61         & 10.51       & 17.36         & 8.18        & 12.81          & 6.23         \\ \hline
Mean error, cm    & 6.45          & 3.69        & 6.33          & 3.29        & 4.13           & 2.19         \\ \hline
RMSE, cm          & 8.09          & 4.61        & 7.85          & 4.03        & 5.15           & 2.70         \\ \hline
Time, sec         & 15.50         & 5.52        & 13.47         & 4.89        & 12.04          & 4.50         \\ \hline
\end{tabular}
\label{tab:error}
\end{table}

The experimental results showed that overall mean error equals 5.6 cm (95\% confidence interval [CI], 4.6 cm to 6.6 cm) for the gesture-drawn trajectories and 3.1 cm (95\% CI, 2.7 cm to 3.5 cm) for the mouse-drawn trajectories. The ANOVA results showed a statistically significant difference between the user's interaction with trajectory patterns: $F$ = 4.006, $p$-value = 0.025 $\textless$ 0.05.
On average, the trajectory generated with gestures deviates by 2.5 cm farther from the “ground truth" path compared to one drawn by a computer mouse. The high positional error has presumably occurred due to the lack of tangible experience during the trajectory generation with gesture interface. This problem could be potentially solved by the integration of a haptic device, which will allow users to feel the displacement of their hand and position it more precisely.

\section{Conclusions and Future Work}

In this paper, a novel swarm control interface is proposed, in which the user leads the swarm by path drawing with the DNN-based gesture recognition and trajectory generation systems. Thus, DronePaint delivers a convenient and intuitive toolkit to the user without any additional devices, achieving both high accuracy and variety of the swarm control. The developed system allowed the participants to achieve high accuracy in trajectory generation (average error by 5.6 cm, max error by 9 cm higher than corresponding values during the mouse input).

In future work, we plan to add an entire body tracking to control a multitude of agents. For example, the movement of the body or hands can change the drone's speed and orientation, which will increase the number of human swarm interaction scenarios. Additionally, we will explore methods to control a swarm in different positioning systems, such as GPS for outdoor performance. Finally, a swarm behavior algorithm for distributing tasks between drones will increase path generation performance and quality.

The proposed DronePaint systems can potentially have a big impact on movie shooting to achieve desirable lighting conditions with the swarm of spotlights controlled by an operator. Additionally, it can be used in a new generation of a light show where each spectator will be able to control the drone display, e.g., navigate the plane, launch the rocket, or even draw the rainbow in the night sky.

\begin{acks}
The reported study was funded by RFBR and CNRS, project number 21-58-15006.
\end{acks}

\bibliographystyle{ACM-Reference-Format}
\bibliography{sample-base}

\end{document}